\documentclass{article}
\usepackage[utf8]{inputenc}
\usepackage{graphicx}
\usepackage{float}
\usepackage{epstopdf}

\title{Estimation of Body Mass Index from Photographs using Deep Convolutional Neural Networks}
\author{Adam Pantanowitz\footnote{Corresponding author: adam -dot- pantanowitz -at- wits.ac.za}, Emmanuel Cohen, \\Philippe Gradidge, Nigel Crowther, \\Vered Aharonson, Benjamin Rosman, David M Rubin\footnote{All authors are with the University of the Witwatersrand, Johannesburg.}\footnote{Preprint}}
\date{August 2019}

\begin{document}

\maketitle
\begin{abstract}
Obesity is an important concern in public health, and Body Mass Index is one of the useful (and proliferant) measures. We use Convolutional Neural Networks to determine Body Mass Index from photographs in a study with 161 participants. Low data, a common problem in medicine, is addressed by reducing the information in the photographs by generating silhouette images. Results present with high correlation when tested on unseen data.
\end{abstract}

\section{Introduction}

Obesity is an important public health concern, and an understanding of the difficulty in reducing obesity in individuals despite the seemingly simple energy balance equation that underlies all weight gain remains elusive~\cite{james2018obesity}.

As with any scientific endeavour, measurement is crucial to achieving insight, and thus a suitable measure of obesity is necessary to advance the research. A number of measures have been proposed including Body Mass Index (BMI), which is defined as the ratio of the mass to the square of the height~\cite{eknoyan2007adolphe}.

There are other measures of obesity which may have merit in terms of refining the measure of metabolic pathology, such as body fat composition, however BMI is arguably one of the easiest measurements to make and remains a reasonable measure of obesity despite some shortcomings~\cite{adab2018bmi}.

An ability to rapidly measure the extent of obesity in populations would constitute a valuable tool for public health research, and the objective of this research is to develop methods to determine BMI from photographic images of subjects.

Work has been done  in determining measures of obesity from  dual-energy x-ray absorptiometry images in 1~554 children \cite{xie2015accurate}, however, this approach is based off measurements gathered invasively, and so does not lend itself to rapid, widespread acquisition of obesity measures.

We present an approach based on individual images taken in a standard setting.  We anticipate that it may be possible to adapt these techniques for large groups of people or individuals who are imaged in non-deal and non-standard settings.

\section{Methodology}

Institutional Review Board approval by the University of the Witwatersrand Human Research Ethics Committee (Medical) on 11 July 2018.

\subsection{Participants}
Anthropometric data and full colour images were gathered from 161 (79 female) Cameroonians and Senegalese participants~\cite{cohen2015development}.

\subsection{Constrained Data}

Machine learning is generally best with large data sets. In this study, however, data are constrained due to the number of participants. The small number of images available made this study particularly challenging. In order to manage this, silhouette images were generated from the original photographs. This substantially reduces the information/complexity of the images.

\subsection{Image Preprocessing and Silhouette Generation}

\begin{figure}
  \centering
  \includegraphics[width=1.0\textwidth]{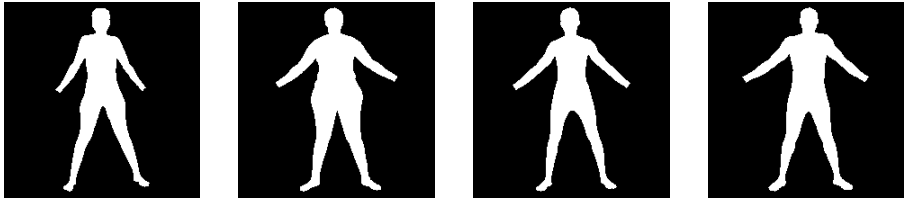}
  \caption{\label{fig:largesil}A random sample of four silhouette images from the 161 study participants.}
\end{figure}

Silhouette images of the 161 participants were manually generated by marking images by hand \textit{s1}, and another set automatically generated \textit{s2} through automated image processing. A sample of the silhouette images is presented in Figure~\ref{fig:largesil}.

Owing to the low count of participants, it is necessary to further reduce complexity by standardising the silhouette images as far as possible, as preprocessing to machine learning.

The images are first standardised in terms of resolution and dimensions using \textit{ImageMagick}, \textit{bash}, and \textit{Python}. The following transforms are applied:

\begin{itemize}
  \item Inversion;
  \item Density conversion;
  \item Resolution standardisation;
  \item Cropping to a rectangular bounding box;
  \item Resizing to a standard width whilst maintaining aspect ratio;
  \item Adding consistent padding; and
  \item Resizing to 64x64 pixels.
\end{itemize}

\textit{s2} is produced by automated processing of the original, full scale images. Several full colour front-facing photographs were taken per participant in the original photographs. Single images were chosen manually to form a subset of images which appeared most consistent in terms of light, and with the least blur.

For both sets, the following techniques implemented in \textit{Python} using \textit{PIL}, \textit{skimage}, and \textit{OpenCV} were used to generate the automatic silhouettes:

\begin{itemize}
  \item Colour transforms;
  \item Distance transforms;
  \item Thresholding;
  \item Morphological opening;
  \item Connected components; and
  \item Region Properties.
\end{itemize}

\begin{figure}
\centering
\includegraphics[width=0.3\textwidth]{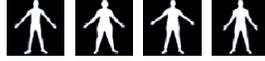}
\caption{\label{fig:smallsil}A random sample of four preprocessed, automatically generated silhouette images from \textit{s2} (slightly scaled, actually 64x64 pixels).}
\end{figure}

This procedure creates a set of silhouette images automatically. The set differs from s1 in that hands and feet silhouettes remain, and a bit of noise around the feet are present due to floor markings remaining as artefacts due to floor markings in the original images. Samples indicating the small silhouette images are presented as \ref{fig:smallsil}. Note that while both sets undergo the same preprocessing to standardise them, the sets differ slightly in their appearance.

\subsection{Deep Learning}
\subsubsection{Constraints in Data}

Deep Learning generally requires large amounts of data. In recent years, image data has been used to generate Convolutional Neural Network (CNN) classification algorithms. Creating binary silhouette images dramatically reduces the information content required for processing.

Data augmentation is implemented with small parameters of rotation (2$\deg$) and width shift ($0.02$) (similar to variations in the original image set), and horizontal image reflections. Augmentations larger than a few degrees of rotation and/or shifts, or augmentations such as height changes, dramatically reduce performance. Data augmentation parameters are optimised using sklearn's Basinhopping algorithm.

Without data augmentation, the model fails to train sufficiently, even with high patience on early stopping.

\begin{figure}[H]
\centering
\includegraphics[width=0.26\textwidth]{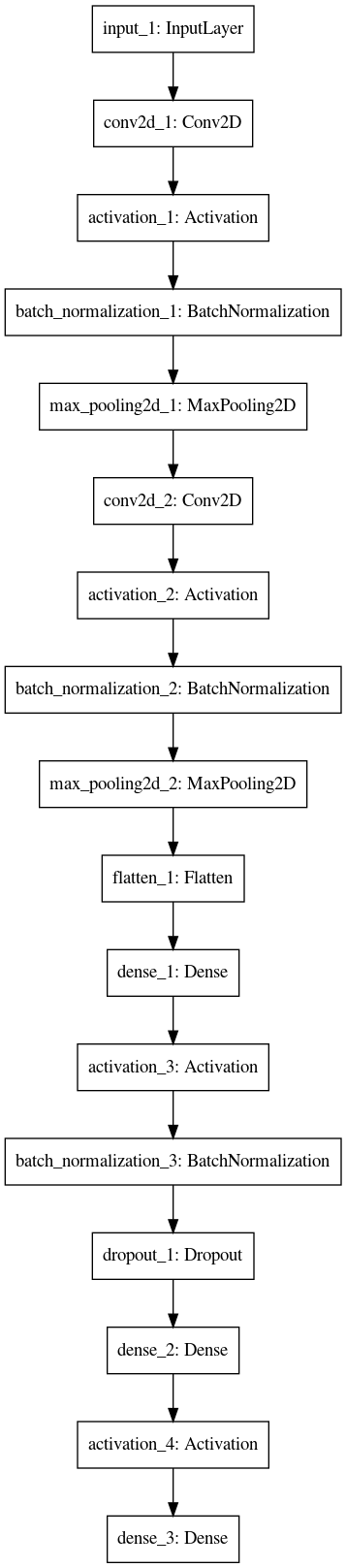}
\caption{\label{fig:model}Exported model architecture.}
\end{figure}

\subsubsection{Convolutional Neural Network}

A two dimensional Convolutional Neural Network (CNN) is implemented in Keras using TensorFlow to perform deep learning on the model. Gridsearch allows for the optimal selection of hyperparameters for training, and the learning and decay rate make substantial difference to whether or not the algorithm converges. The optimal parameters are found to be using the Adam algorithm with a learning rate of $0.02$, and decay of $1\times10^{-4}$.

Data are split in a few variations, so as to optimise the trade-off between each set. The minimum training size with reasonable performance is with approximately 80 images for training, and the remainder making up the validation set. Optimisation is performed to try maximise the test set while maintaining a validation set sufficient to early stop the training of the model.

Training is performed using source images and the target of normalised BMI between zero and one.

The CNN architecture and model structure export is presented in figure~\ref{fig:model}. A few architectures were tweaked and evaluated, but sufficient performance was achieved and model structure search was therefore non-exhaustive.

\section{Results}

The system allows BMI to be predicted with a linear regression of above 0.96 for most runs. Sufficiently low validation performance is a signal of model convergence, and robust prediction results follow.

\begin{figure}[H]
\centering
\includegraphics[width=1.0\textwidth]{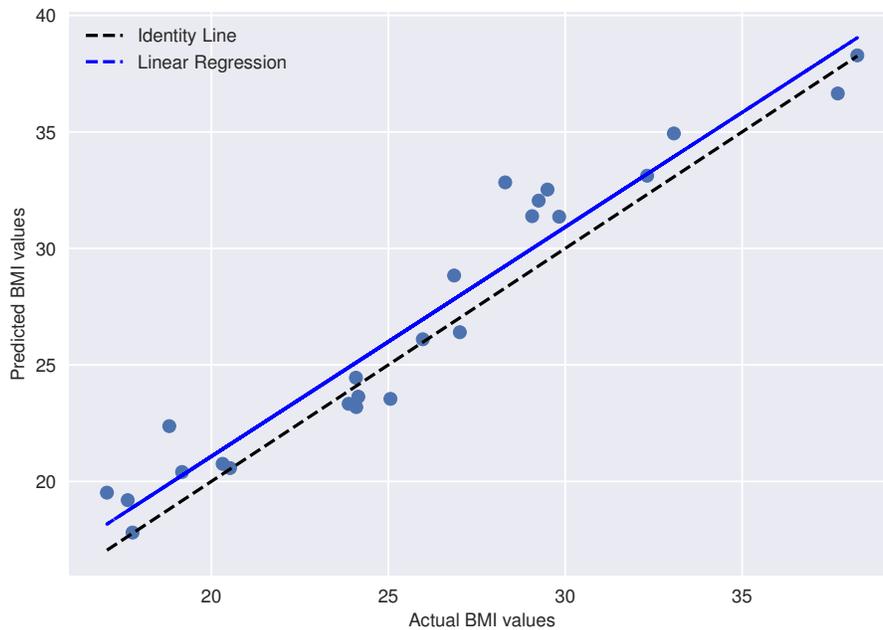}
\caption{\label{fig:results}Scatter Plot of Predicted vs Actual BMI for unseen data.}
\end{figure}

Figure~\ref{fig:results} presents the predictive performance of the model.

\section{Conclusion}

We successfully implement a system to predict BMI from a small data set of photographic images of participants. This solution is toward a possible public health screening tool to assist in supporting/measuring health initiatives in areas of endemic obesity or malnutrition. If used in this context, as the system makes use of silhouette imagery, there is an automatic benefit of the system creating anonymity, as the silhouette imagery is inherently de-identified.

\bibliography{main}
\bibliographystyle{ieeetr}

\end{document}